\pdfoutput=1

\documentclass[11pt]{article}

\usepackage{emnlp2021}

\usepackage{times}
\usepackage{latexsym}

\usepackage[T1]{fontenc}

\usepackage[utf8]{inputenc}

\usepackage{microtype}

\usepackage{bm}
\newcommand{\argmax}{\mathop{\mathrm{argmax}}}

\usepackage{graphicx}
\usepackage{algorithmic}
\usepackage{algorithm}
\usepackage{amsmath}
\usepackage[multiple]{footmisc}
\title{Simultaneous Neural Machine Translation with Constituent Label Prediction}

\author{Yasumasa Kano$^{1}$ \quad Katsuhito Sudoh$^{1,2}$ \quad Satoshi Nakamura$^{1,2}$\\
  $^{1}$Nara Institute of Science and Technology (NAIST), Japan \\
  $^{2}$Center for Advanced Intelligence Project (AIP), RIKEN, Japan\\
  \texttt{\{kano.yasumasa.kw4, sudoh, s-nakamura\}@is.naist.jp}}

\begin{document}
\maketitle
\begin{abstract}

Simultaneous translation is a task in which translation begins before the speaker has finished speaking, so it is important to decide when to start the translation process. However, deciding whether to read more input words or start to translate is difficult for language pairs with different word orders such as English and Japanese. Motivated by the concept of pre-reordering, we propose a couple of simple decision rules using the label of the next constituent predicted by incremental constituent label prediction. In experiments on English-to-Japanese simultaneous translation, the proposed method outperformed baselines in the quality-latency trade-off.

\end{abstract}

\section{Introduction}
\label{introduction}
\begin{table*}[t]
\centering
\begin{tabular}
{p{0.30\textwidth}p{0.40\textwidth}}
\hline
Source sentence   & I \textbf{bought} \emph{ a pen}.\\
Monotonic translation &  watashi wa \textbf{katta} \emph{pen wo}.\\
Full-sentence translation & watashi wa \emph{pen wo} \textbf{katta}.\\
\hline
\end{tabular}
\caption{Translation from English (SVO) to Japanese (SOV)}
\label{tab:svotosov}
\end{table*}

\begin{table*}[t]
\centering
\begin{tabular}
{p{0.30\textwidth}p{0.40\textwidth}}
\hline
Boundary prediction   & I / \textbf{bought} \emph{a pen}.\\
Simultaneous translation & watashi wa / \emph{pen wo} \textbf{katta}.\\
\hline
\end{tabular}
\caption{Example of English-to-Japanese translation using proposed method with segment-boundary prediction}
\label{tab:proposedsegment}
\end{table*}
Simultaneous machine translation is a task in which the machine starts outputting a translation before reading the entire input sentence. This task is more difficult than full-sentence translation because it translates the initial part of a sentence without the context of the latter part. This involves a trade-off between delay and quality of the translation; using a longer context should improve translation quality at the cost of a longer delay, and vice versa. In practice, we should control the latency so that it's not too large, but we may also need to allow a long latency depending on the situation.

Most of the recent simultaneous translation models \cite{ma-etal-2019-stacl,arivazhagan-etal-2019-monotonic,raffel-2017,arivazhagan-etal-2019-monotonic,xutaima-2020,dalvi-etal-2018-incremental,gu-etal-2017-learning,alinejad-etal-2018-prediction,cho-2016,zheng-etal-2020-simultaneous,zheng-etal-2019-simpler,zhang-etal-2020-learning-adaptive} are based on neural machine translation (NMT), although earlier studies were based on statistical machine translation \cite{rangarajan-sridhar-etal-2013-segmentation,grissom-ii-etal-2014-dont,oda-etal-2014-optimizing,oda-etal-2015-syntax}.
In simultaneous NMT, there are two major approaches: those in which a latency hyperparameter is given before the training and those in which it is given at the time of inference. 

The former approach requires training a model individually for each pre-defined latency setting, while the latter approach uses a single model for different latency conditions.
Most human simultaneous interpreters would not need such long training to slightly adjust latency, while it takes much more time to learn other languages to develop their translation skill.
Therefore, the latter approach is closer to the learning process of human simultaneous interpreters.

wait-k \cite{ma-etal-2019-stacl} is a simple simultaneous NMT method of the former approach that waits \textit{k} tokens before starting to translate. It also has variants within the latter approach called test-time wait-k, in which k is determined at the inference time. wait-k had better performance than test-time wait-k in that study's experiments. 

There is another method in the latter approach that uses Meaningful Unit \cite{zhang-etal-2020-learning-adaptive}. In this model, chunk-based incremental decoding is done at inference time by segmentation with a boundary predictor. This model outperformed baselines of the former approach. They refined their basic boundary predictor to deal with sentence pairs in which full-sentence translation needs long-distance reordering. However, its training process is very complicated: It first generates monotonic translations, fine-tunes the NMT model with them, then generates an oracle boundary with the model, and finally fine-tunes a boundary-prediction model based on BERT \cite{devlin-etal-2019-bert}.

Simultaneous translation is still difficult for language pairs such as English-Japanese, which often require long-distance reordering.
To tackle the reordering problem, we propose an input-segmentation method for simultaneous translation, using a couple of simple rules and incremental prediction of the label of a syntactic constituent coming immediately after the input existing so far.
This is not dependent on the trained NMT. Therefore, once we create it, it is reusable for other models. 

Our proposed method is inspired by Head Finalization \cite{isozaki-etal-2010-head}.
Head Finalization reorders words of the source sentence before translating from an SVO (Subject-Verb-Object) language to an SOV language in full-sentence statistical machine translation.
This method moves a syntactic head into a later position so that the word order of the source language (e.g., English) becomes similar to that of the target language (e.g., Japanese).
This enables us to monotonically translate from English, which is a typical SVO language, to Japanese, a typical SOV language. 

Recent NMT models like Transformer \cite{vaswani-2017} works well on reordering in general, so this kind of pre-reordering is not usually used. However, simultaneous translation monotonically reads input words one by one, and therefore the difference in word order remains a problem. As shown in Table~\ref{tab:svotosov}, monotonic translation often becomes unnatural compared to full-sentence translation. The part ``bought a pen'' should be translated to \emph{pen wo katta} by reversing the word order. Therefore, after reading the word ``bought,'' it is important to wait for future words without starting to translate it. In this case, ``I'' is the last word that does not require reordering. This word is regarded as a segment boundary to start a partial translation.

Table~\ref{tab:proposedsegment} shows an example of our proposed segmentation.
Suppose we predict the next constituent label as a verb phrase (VP) after reading an input word ``I.''
This shows the possibility that the next words should be reordered, so the ``I'' becomes the boundary.
Once detecting the boundary, NMT model starts to translate ``I'' into ``watashi wa.''
After that, the model restarts to read the remaining input words, then translates ``bought a pen'' into ``pen wo katta.''
The total output of simultaneous translation based on the proposed segmentation is the same as that of full-translation in this simple example.
By ensuring that the Verb and its Object of the source sentence are included in a single segment, it is possible to output translation while maintaining the SOV-like structure of the target language.

In experiments on English-to-Japanese simultaneous translation, the proposed method outperformed baselines in the quality-latency trade-off.

\section{Related work}
In statistical machine translation, there are several approaches to finding boundaries of segments for simultaneous translation. \citet{oda-etal-2014-optimizing} proposed a method to choose segment boundaries that maximize the BLEU score. \citet{rangarajan-sridhar-etal-2013-segmentation} proposed segmentation strategies based on lexical cues.

In NMT, there have been many studies on simultaneous translation. The amount of latency is decided either before training or at inference time. wait-k \cite{ma-etal-2019-stacl} is the simplest variant using fixed latency: It simply waits for k tokens before starting translation \cite{ma-etal-2019-stacl}.
The latency policy can be learned from a parallel corpus together with an NMT model. MILk \cite{arivazhagan-etal-2019-monotonic} and other approaches \cite{raffel-2017,xutaima-2020} used a latency-augmented loss function in training to balance latency and accuracy.

In contrast, the latency policy can be learned with a pre-trained NMT model, such as test-time wait-k \cite{ma-etal-2019-stacl} and STATIC-RW \cite{dalvi-etal-2018-incremental}.
These have fixed policies that wait for the fixed number of tokens before translation, but there are other models that learn a more flexible policy for a given pre-trained NMT model.
Some studies use reinforcement learning to learn an adaptive READ/WRITE policy \cite{grissom-ii-etal-2014-dont,satija-2016,gu-etal-2017-learning,alinejad-etal-2018-prediction}.
Training by reinforcement learning can be unstable depending on the condition.
One method that does not use reinforcement learning is wait-if-* \cite{cho-2016}, which translates and segments jointly to maximize the translation quality.
\citet{zheng-etal-2020-simultaneous} extended wait-k to an adaptive policy by adaptively choosing the strategy at inference.
There is another method that generates oracle READ/WRITE actions by a pre-trained NMT model and predicts actions using a neural network model \cite{zheng-etal-2019-simpler}.
Meaningful Unit \cite{zhang-etal-2020-learning-adaptive} works along the same lines and has outperformed baselines such as MILk and wait-k. 

With respect to the use of syntactic clues for simultaneous translation, \citet{oda-etal-2015-syntax} proposed a method to incrementally parse an incomplete sentence by predicting unseen syntactic constituents on the right and left side of each segment.
They concatenated the predicted constituents and the words in a segment and then input the result into tree2string translation.
They decided to wait for more tokens or output the translation depending on where the constituents appear in the translation result. 

Our proposed method is based on chunk-based simultaneous translation using chunk boundary detection with simple rules on next-constituent labels.
It basically segments an input before a verb phrase. This is much simpler and easier to implement than the work by \citet{zhang-etal-2020-learning-adaptive} and \citet{oda-etal-2015-syntax}.

\section{Proposed Method}
\begin{figure*}[t]
 \centering
 \centerline{\includegraphics[width=16cm]{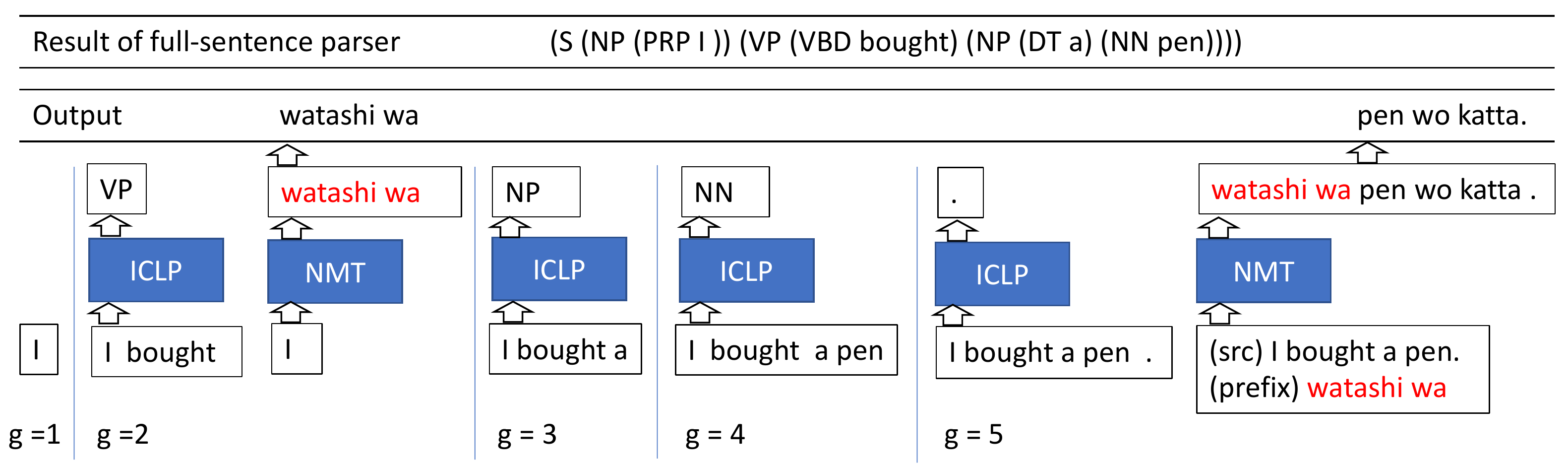}}
 \vspace{-0.2cm}
 \caption{One look-ahead ICLP gives constituent labels. When a boundary is detected based on the label and rules, NMT starts to translate the source subsequence. The previous translation, which is red in the figure, is used as prefix words for the next translation. EOS is omitted for simplicity in the figure.}
\label{fig:overview}
\end{figure*}

\begin{table*}[t]
\centering
\begin{tabular}{lcccccccccc}
\hline
\textbf{segmentation} & You & / can & save & time & by & / doing &  this & . \\
\textbf{constituent label} &  & VP  & VP   & NP  & PP & S    &  NP    & . \\
\hline
\textbf{syntax tree} &  \multicolumn{8}{l}{(S (NP (PRP You)) (VP (MD can) (VP (VB save) (NP (NN time))}\\
& \multicolumn{8}{l}{(PP (IN by) (S (VP (VBG doing) (NP (DT this))))))) (. .))} \\
\hline
\end{tabular}
\caption{Example result of one look-ahead ICLP with a minimum segment size of one.}
\label{tb:willexample}
\end{table*}

Figure~\ref{fig:overview} shows a step-by-step example of our proposed method
described in this section.

\subsection{Standard Simultaneous Translation}
A standard NMT for full sentences is represented by the following equation:
\begin{equation}
p_{full}(Y|X) = \prod_{t=1}^{|Y|} P(y_t | X, y_{<t}),
\end{equation}
where $X = x_1,x_2,...,x_n$ is an input sentence consisting of $n$ tokens and $Y = y_1,y_2,...,y_m$ is a predicted target language sentence consisting of $m$ tokens.

A simultaneous NMT uses only a prefix of the input to predict a target language token:
\begin{equation}
p_{simul}(Y|X) = \prod_{t=1}^{|Y|} P(y_t | x_{g(t)},y_{<t}),
\end{equation}
where $g(t)$ is a monotonic non-decreasing function representing the number of read source tokens to output the $t$th target token. 

\subsection{Chunk-based Simultaneous Translation}
We use chunk-based incremental decoding for our simultaneous translation model and
a full-sentence NMT model trained in a standard manner.
However, at the time of inference, we translate the current prefix upon chunk segmentation while keeping the previously translated output unchanged.

Suppose we have already translated input chunks $\bm{X}^{i-1} = X_1, X_2, ..., X_{i-1}$ into an output prefix also represented by chunks: $\widetilde{\bm{Y}}^{i-1} = \widetilde{Y}_{1}, \widetilde{Y}_{2}, ..., \widetilde{Y}_{i-1}$, while translating the next input chunk $X_i$ into $\widetilde{Y}_{i}$.
We restart the translation from the beginning using all of the available input chunks $X_{1}^{i}$.
This is similar to an approach called \emph{re-translation} that generates translations from scratch for every new input word \cite{Niehues+2016,arivazhagan-etal-2020-translation},
but we apply forced decoding to $\widetilde{\bm{Y}}^{i-1}$ in the output prefix.
The probability of the prefix $\widetilde{\bm{Y}}^{i}$ can be denoted as follows:
\begin{multline}
    p_{prefix} (\widetilde{\bm{Y}}^{i} | \bm{X}^{i}) = \\
    p_{full} (\widetilde{\bm{Y}}^{i-1} | \bm{X}^{i}) \times p_{chunk} (\widetilde{Y}_{i} | \bm{X}^{i}, \widetilde{\bm{Y}}^{i-1}).
\end{multline}
The first term is calculated in the same way as the standard full-sentence NMT in Eq. (1) through forced decoding, and the second term is decomposed as follows, letting $\widetilde{Y}_{i} = y_1^i, y_2^i, ..., y_{|\widetilde{Y}_i|}^i$:
\begin{equation}
    p_{chunk} (\widetilde{Y}_{i} | \bm{X}^{i}, \widetilde{\bm{Y}}^{i-1}) = \prod_{t=1}^{|\widetilde{Y}_i|} P(y_t^i | \bm{X}^{i}, \widetilde{\bm{Y}}^{i-1}, y_{<t}^i).
\end{equation}

This can be more efficient than an incremental Transformer \cite{ma-etal-2019-stacl} that refreshes the encoder for every input word, since our chunk-based translation refreshes the encoder for every input \emph{chunk}, which usually consists of multiple words.

\subsection{Chunk Segmentation}
We use constituent labels for our rule-based chunk segmentation as follows.

\subsubsection{Incremental Constituent Label Prediction}
We predict the label of a syntactic constituent coming after a sentence prefix at the current time-step.
We call this process \emph{Incremental Constituent Label Prediction} (ICLP).
Here, we define this \emph{next constituent} as the one coming next to the sentence prefix in pre-order tree traversal.
However, this label prediction is not easy without observations on the next constituent.
In this work, we allow one look-ahead, where
we read one more word and predict the label of the constituent starting from that word.
This causes an additional delay by one word but improves the prediction accuracy.
Suppose we have an input sequence $W = [w_1, w_2, ..., w_{|W|}]$.
The one look-ahead ICLP predicts the constituent label $c_i$ upon the observation of $w_i$, as follows:
\begin{equation}
    c_i = \argmax_{c' \in C} p (c' | w_{\leq i}),
\end{equation}
where $C$ is a set of constituent labels.
Only a prefix word subsequence is fed into the ICLP, so previous label predictions do not affect later ones.

We can train the ICLP model as a multi-class classifier using a set of training instances in the form of prefix-label pairs.
One sentence generates several instances for training data:
$(w_1, c_1)$, $(w_1, w_2, c_2)$, $(w_1, w_2, w_3, c_3)$, $(w_1, w_2, w_3, w_4, c_4)$, and so on. 
We implemented the ICLP model in two different ways using LSTM \cite{hochreiter-1997} and BERT \cite{devlin-etal-2019-bert}.

\subsubsection{Segmentation Rules}
Table~\ref{tb:willexample} shows an example of a result by the one look-ahead ICLP.
We use one basic and two supplemental rules for chunk segmentation as follows.
\begin{itemize}
  \item Segment the input coming just before constituents labeled \texttt{S} and \texttt{VP}.
  \item If the previous label is \texttt{S} or \texttt{VP}, do not segment the input.
  \item If the chunk is shorter than the minimum length, do not segment the input.
\end{itemize}

In incremental translation from \emph{Subject-Verb-Object} to \emph{Subject-Object-Verb}, the subject can be translated before observing the verb coming next, but the verb should be translated after observing the object.
Therefore, the chunk boundary should be between the subject and verb, not between verb and object.
To achieve this, we employ a simple rule to segment a chunk just before \texttt{VP}.
We also include \texttt{S} in the rule just as with \texttt{VP} because \texttt{S} (simple declarative clause) often appears in the form of a unary branch ``(\texttt{S} (\texttt{VP} ...))'' as shown in Table~\ref{tb:willexample}.

However, in cases such as ``can save'' in the example, \texttt{VP} occurs again immediately after the segmentation before ``can.''
The basic rule suggests segmentation before ``save,'' but it does not seem appropriate.
Therefore, we introduce the minimum segment size to avoid such over-segmentation as a hyperparameter to control the accuracy-latency trade-off.
If the hyperparameter is larger than one, the chunk segmentation after ``You'' in the example does not occur.

\section{Experimental setup}
\subsection{Dataset and preprocessing}

We conducted experiments on English-Japanese (En-Ja) translation. We also tried English-German (En-De) translation to investigate the difference in language pairs. 

For En-Ja, the model was trained on 17.9 M sentence pairs from WMT2020 and  fine-tuned on 223 K sentence pairs from IWSLT2017. We used 5312 sentence pairs for the development set from  dev2010, tst2011, tst2012, and tst2013 of IWSLT. We evaluated the model on  1442 sentence pairs from dev2021 of IWSLT. 

For En-De, the model was trained on 4.5 M sentence pairs from WMT2014 and  fine-tuned on 206 K sentence pairs from IWSLT2017. We used 5589 sentence pairs for the development set from  dev2010, tst2011, tst2012, and tst2013 of IWSLT. We evaluated the model on  1,080 sentence pairs from tst2015 of IWSLT. 

We tokenized English and German sentences with \texttt{tokenizer.perl} in Moses \cite{koehn-etal-2007-moses} and Japanese sentences with MeCab \cite{kudo2005}.
For each language pair, we used subwords based on Byte Pair Encoding (BPE) \cite{sennrich-etal-2016-neural} with a shared vocabulary of 16 K entries.
To develop the subword vocabulary, we used all of the in-domain training sentences (IWSLT) and one million out-of-domain sentences (WMT).

We trained the ICLP models using Penn Treebank 3 \cite{marcus-etal-1993-building} for training, excluding a randomly selected one percent of sentences reserved for the development set.
We used NAIST-NTT TED Talk Treebank \cite{neubig14iwslt} for the evaluation set. The number of training, development, and test instances (e.g., the number of labels to be predicted) were 2.8 M, 27.9 K, and 21.9 K, respectively. Note that multiple ICLP instances are induced from what a single parse tree generates.

\subsection{Model settings}
We compared the following four models.  All of them were based on the Transformer-base \cite{vaswani-2017}. 

\begin{description}
\item[wait-k]\mbox{}\\
The range of \texttt{k} is [2, 4, 6,..., 30].
\item[Meaningful Unit]\mbox{}\\
The hyperparameter is \texttt{p}, which is the threshold of the probability of a boundary.
The ranges of \texttt{p} are [0.5, 0.1, 0.15,..., 0.95], [0.99, 0.991, 0.992,..., 0.999], and [0.9991, 0.9992,..., 0.9999].
Monotonic translation of Meaningful Unit was generated from the fine-tuning dataset by the fine-tuned NMT model.
We used their refined Meaningful Unit method, which improved the translation quality at low latency \cite{zhang-etal-2020-learning-adaptive}\footnote{\citet{zhang-etal-2020-learning-adaptive} removed the monotonic translations with a lower score than full-sentence translation.
However, it is rare for a monotonic translation to have a higher score than full-sentence translation.
Consequently, few sentences remained in our setting.
Therefore, we improved the translation quality by preventing over-translation instead of removing it.
Once the same words are output four times continuously or the target length becomes four times longer than the source length, we expand the source prefix.}.
They used a two look-ahead boundary predictor in their experiments.
We additionally tried a one look-ahead predictor because it is not certain how many future words should be used for the predictor. 
\item[Fixed-size segmentation]\mbox{}\
This simply segments an input with a fixed length specified by a hyperparameter \texttt{f}, which means the boundary comes every \texttt{f} subwords or words.
The range of \texttt{f} is [2, 4, 6,..., 30] for words and [4, 8, 12,..., 60] for subwords.
\item[ICLP]\mbox{}\\
The hyperparameter is \texttt{m}, which means the minimum number of words in one segment. The range of \texttt{m} is [1, 2, 3, …, 29]. 
\end{description}

We controlled hyperparameters to adapt to a wide range of  latency.
The hyperparameter is given both in the training and at the inference time for wait-k, but it is given only at the inference time for other models.
Therefore, we trained the wait-k model for each \texttt{k} while in other approaches a single NMT model is commonly used.

We used fairseq \cite{ott-etal-2019-fairseq} to implement these models and basically followed the official baseline for IWSLT 2021\footnote{\url{https://github.com/pytorch/fairseq/blob/master/examples/simultaneous_translation/docs/enja-waitk.md}}\footnote{\url{https://github.com/pytorch/fairseq/issues/346}} to set the hyperparameters. We saved checkpoints every 5000 updates for pre-training and every 200 updates for fine-tuning.
Other hyperparameters were the same for pre-training and fine-tuning.
We stopped training early with patience 4.
The max-tokens for the mini batch size was  4096, and weights were updated every 4 mini batches. We set the learning rate to 0.0007 and trained the model on a single GPU. The last three models used the same NMT model.
We used beam search within chunks in a standard way and chose 1-best hypotheses at the end of chunk translation.
The beam size was four for the chunk-based and full-sentence models.
We used greedy decoding for wait-k.

We implemented two types of ICLP models as mentioned earlier.
For the LSTM-based ICLP, we used two-layered unidirectional LSTMs to encode an input sentence with a fully connected layer for the constituent label prediction.
The numbers of dimensions for embedding and hidden states are 512.
We tokenized English sentences using \texttt{tokenizer.perl} in Moses and Byte Pair Encoding \cite{sennrich-etal-2016-neural} with a vocabulary of 16 K entries.
For the BERT-based ICLP, we used a BERT-based classifier with an additional fully connected layer over the \texttt{[CLS]} token,
implemented using Huggingface transformers \cite{wolf-etal-2020-transformers} with a pre-trained model \texttt{bert-base-uncased} and the corresponding subword tokenizer.
For both models, the input was a subword sequence, so the constituent label prediction was made upon the observation of an \emph{end-of-word} subword.
The following training conditions were commonly applied to both models: learning rate of 5e-5, training batch size of 512 instances, checkpoints saved at the end of every epoch, and early stopping with the patience of three epochs.

\subsection{Evaluation}
We used SimulEval \cite{ma-etal-2020-simuleval} to evaluate the quality and latency of simultaneous translation. BLEU \cite{papineni-etal-2002-bleu} was used to evaluate quality. We used Average Lagging (AL) \cite{ma-etal-2019-stacl} to evaluate the latency. AL is widely used and defined by the following equation:
\begin{equation}
AL_g(X, Y) = \frac{1}{\tau_g(|X|)}\sum_{t = 1}^{\tau_g(|X|)}g(t) - \frac{t-1}{\gamma}.
\end{equation}
$\tau_g(|X|)$ is the decoding step when the source sentence finishes. It counts latency up to the $\tau_g(|X|)$ th target token predicted just after reading the final source token.  $\gamma$ is defined as $|Y|/|X|$. When the source length $|X|$ equals target length $|Y|$, AL of wait-k equals its k. 
In this experiment, the latency was calculated on character level for En-Ja, and word level for En-De.

\subsection{Results}
\begin{figure}[t]
 \centering
 \centerline{\includegraphics[width=8.5cm]{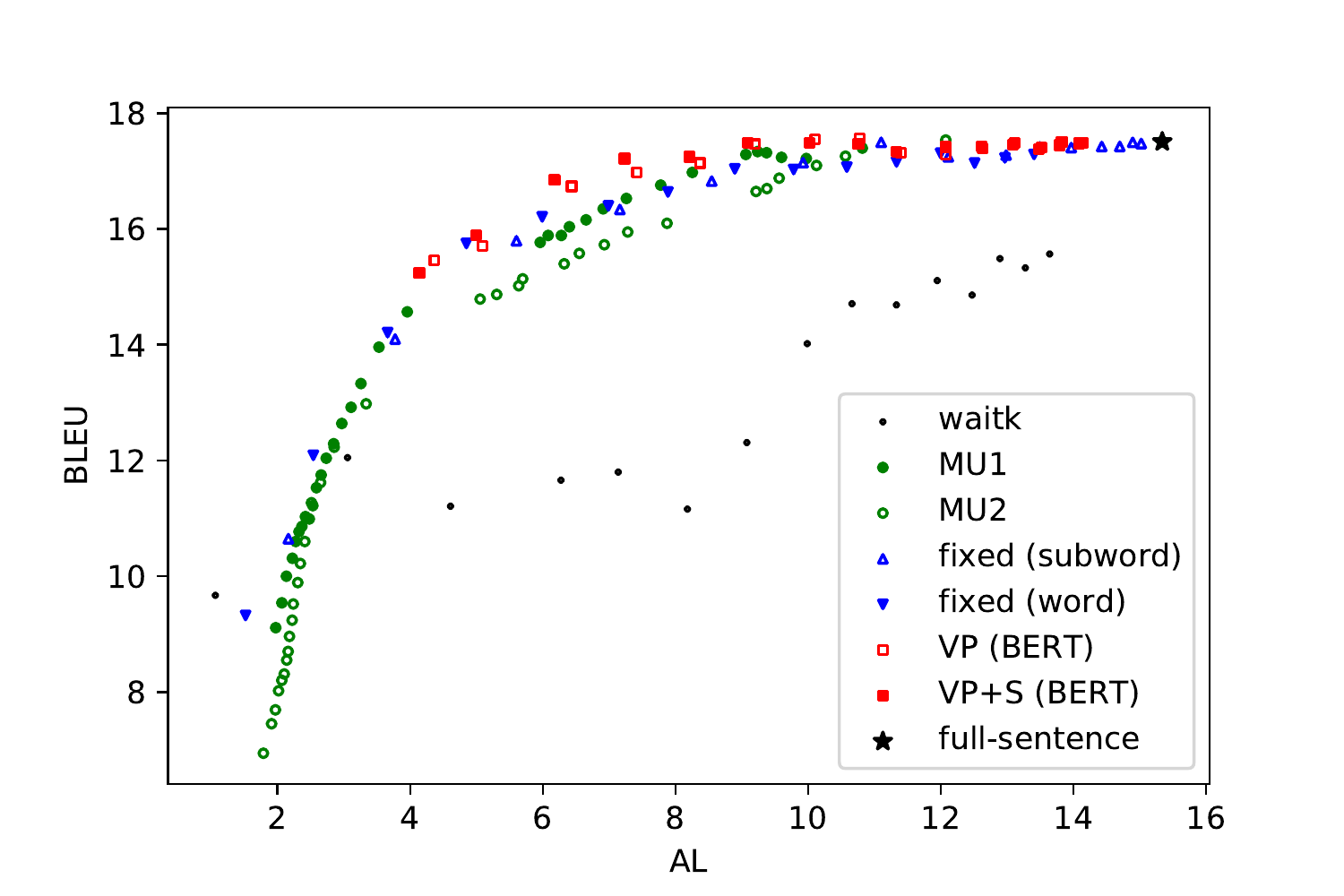}}
 \caption{Scatter plot of BLEU and AL (En-Ja).
 MU1 and MU2 correspond to Meaningful Unit with one and two look-ahead respectively.}
\label{fig:result_all}
\end{figure}

\begin{figure}[t]
 \centering
 \centerline{\includegraphics[width=8.5cm]{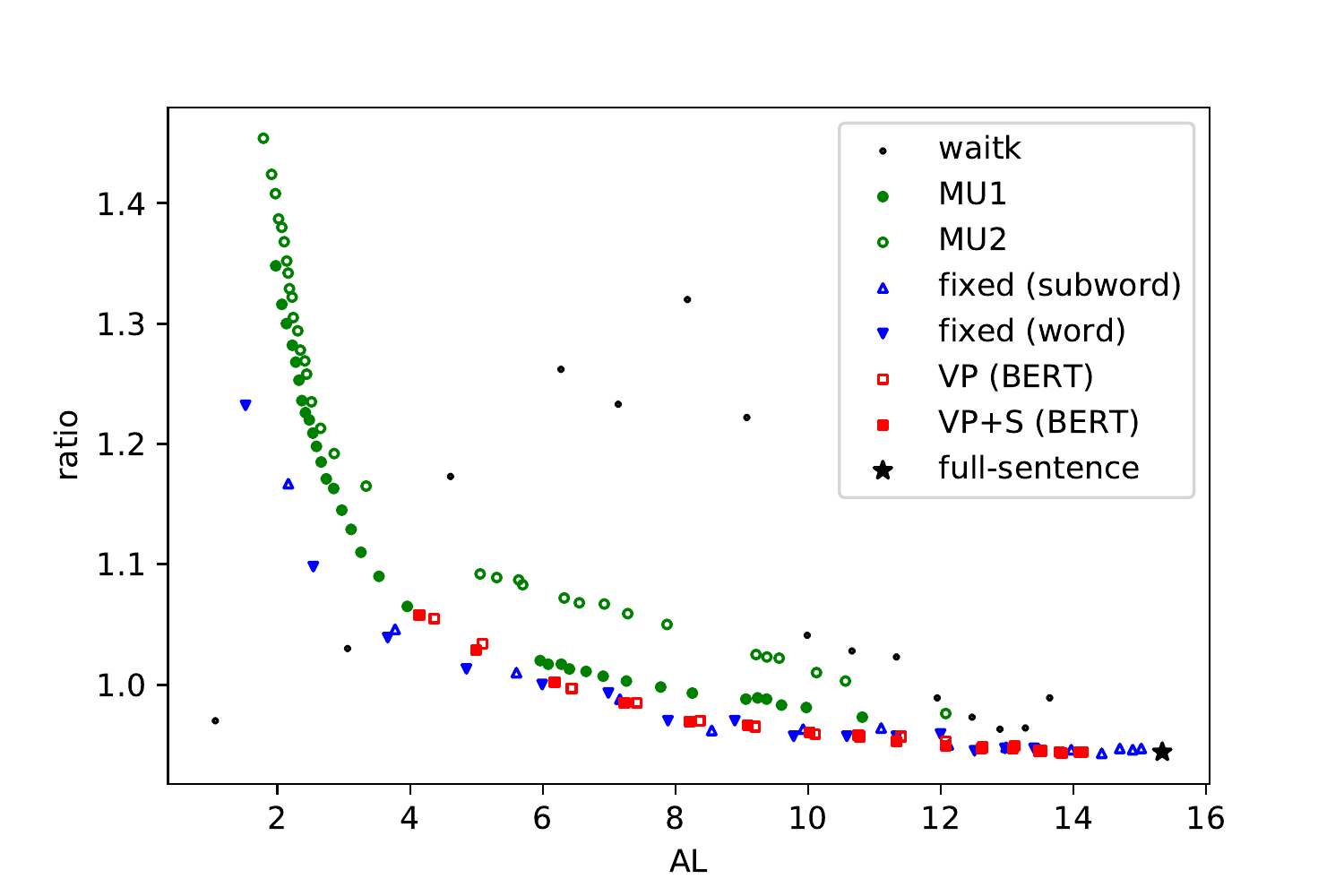}}
 \caption{Scatter plot of length ratio and AL (En-Ja)}
\label{fig:result_ratio}
\end{figure}
We illustrate the results of English-Japanese translation in Figure~\ref{fig:result_all}. Our proposed method outperformed baselines in a wide range of AL.
Most of the points of the proposed method appear to the upper-left of the other methods, thus showing the best performance.
We compared the use of segmentation rules based on \texttt{VP} and \texttt{VP}+\texttt{S}.
The points shifted to the left by adding \texttt{S} as boundary because it increased the number of boundaries and decreased latency.
Although we tried the different look-ahead lengths of one and two for the boundary predictor of Meaningful Unit, our proposed model outperformed both of these models in a wide range of latency.

The difference between wait-k and the models using the full-sentence translation model was large in the quality-latency trade-off.
Surprisingly, the fixed-size segmentation was also effective. When the segment size was fixed, it did not make a large difference in the result, regardless of whether the unit was a subword or a word.

\section{Analysis}
\subsection{Length ratio}
Figure~\ref{fig:result_ratio} shows the length ratios of translation hypotheses and references with different latency parameters. Too large a ratio decreases the BLEU score and makes the content delivery difficult both in text (subtitles) and speech (text-to-speech).

The length ratio of wait-k was unstable compared to other models because it was trained individually for each k. 

Except for wait-k, the length ratios were large in the range of small latency, probably due to the condition mismatch between training and inference. These NMT models were trained on full sentences, but they were used to translate short segments in the inference. Therefore, they tend to output longer segment translations than expected. Their ratios gradually decrease as AL increases and the length of segments becomes closer to the length of full sentences.

\begin{figure}[t]
 \centering
 \centerline{\includegraphics[width=8.5cm]{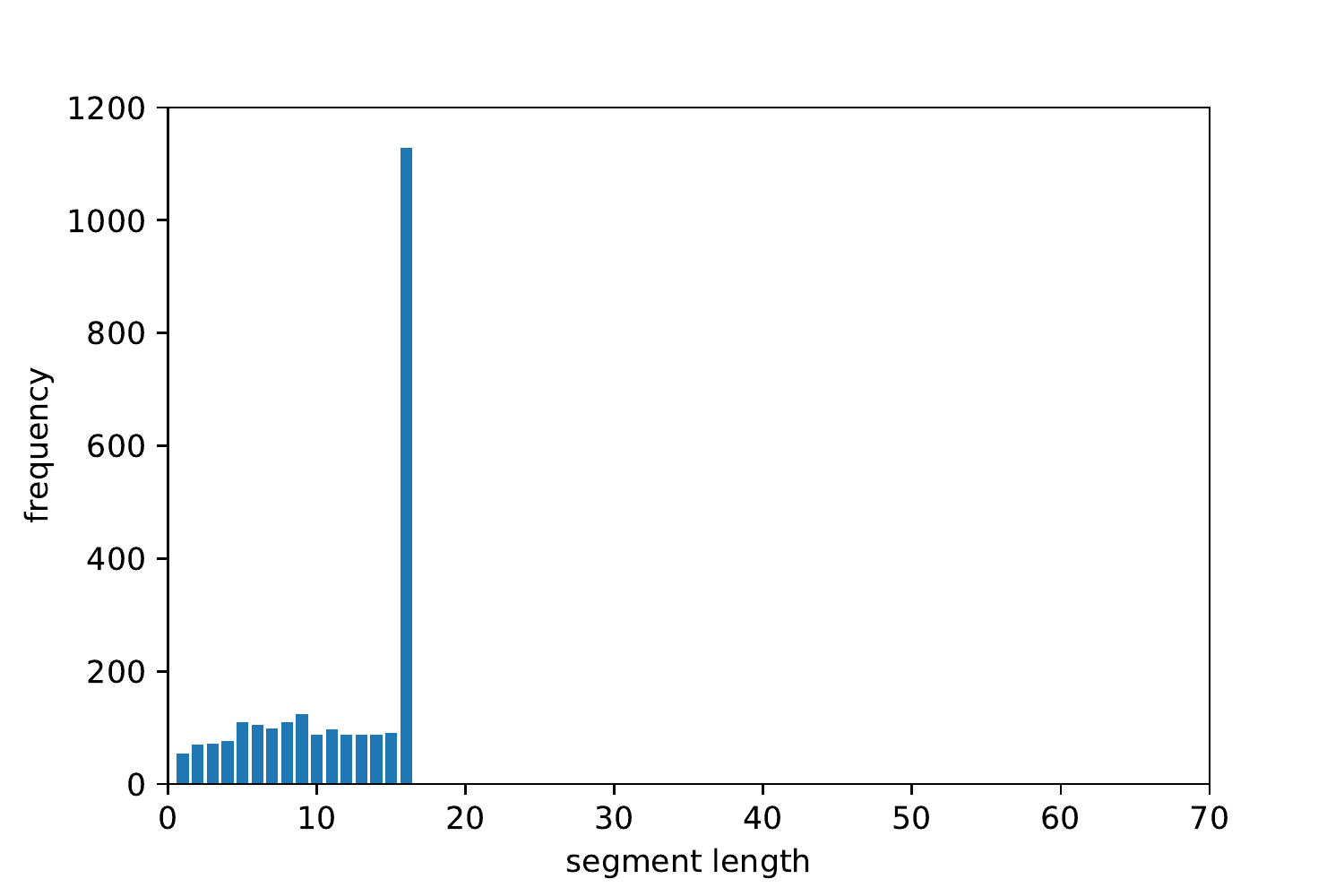}}
 \caption{Segment length distribution of fixed-size segmentation with 16 subwords for AL 7.16 (En-Ja test)}
\label{fig:distfixed}
\end{figure}

\begin{figure}[t]
 \centering
 \centerline{\includegraphics[width=8.5cm]{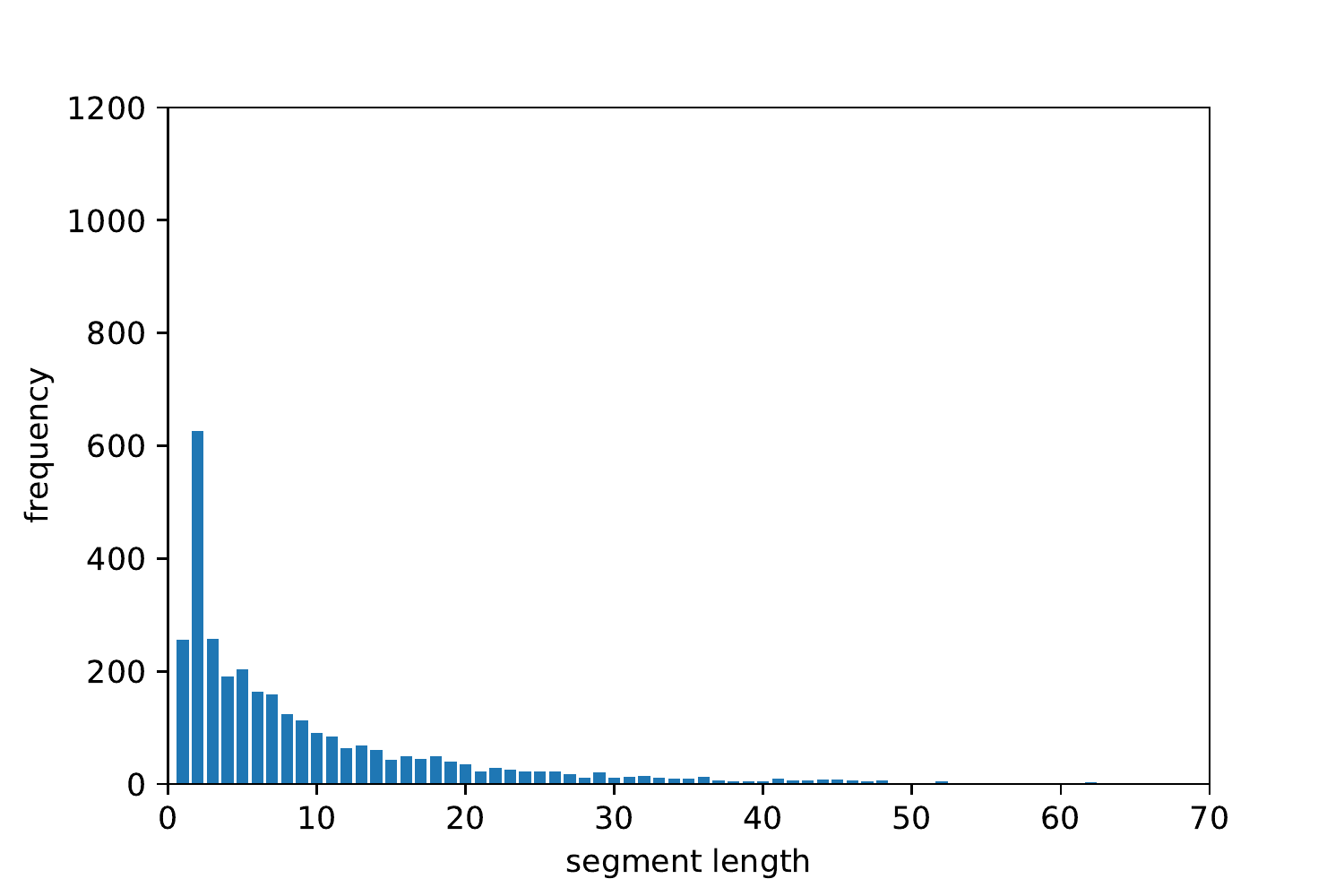}}
 \caption{Segment length distribution of one look-ahead Meaningful Unit for AL 7.26 (En-Ja test)}
\label{fig:distMU}
\end{figure}
\begin{figure}[t]

 \centering
 \centerline{\includegraphics[width=8.5cm]{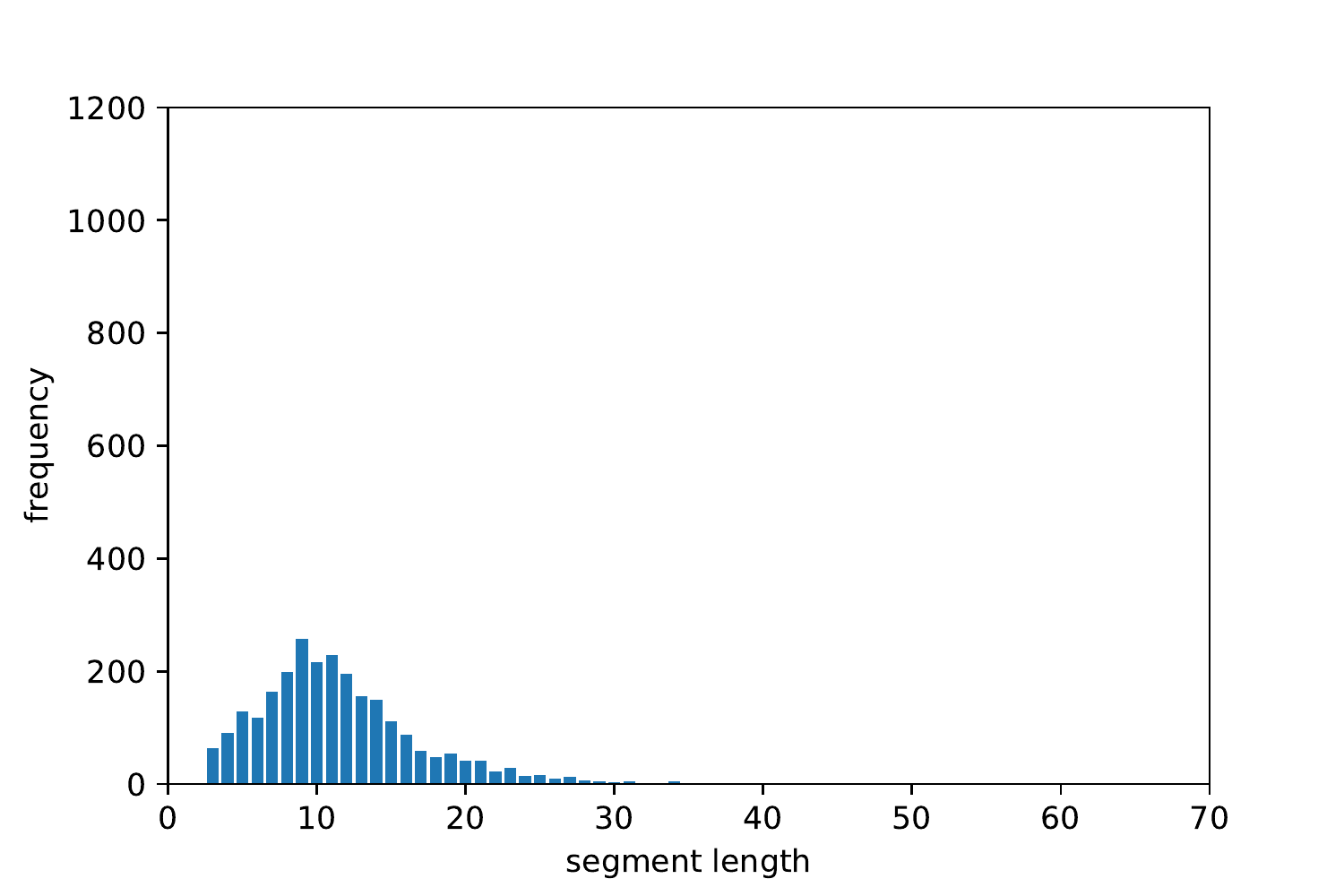}}
 \caption{Segment length distribution of one look-ahead ICLP model dividing with label \texttt{S} and \texttt{VP} for AL 7.23 (En-Ja test)}
\label{fig:distICLP}
\end{figure}

\begin{table}[t]
\centering
\begin{tabular}{lcc}
\hline
\textbf{Label}  & \textbf{AL} & \textbf{BLEU}   \\
\hline
\textbf{Fixed (16 subwords)} & 7.16 &  16.34 \\
\textbf{1 look-ahead MU} & 7.26 & 16.53 \\
\textbf{1 look-ahead ICLP (VP+S)} & 7.23 & 17.22 \\
\hline
\end{tabular}
\caption{BLEU results for AL close to 7}
\label{tb:dist}
\end{table}

\subsection{Segment length distribution}

Figures~\ref{fig:distfixed},\ref{fig:distMU} and \ref{fig:distICLP} show the distributions of source segment length in the En-Ja test set for which AL is close to 7.2.
Table~\ref{tb:dist} shows their corresponding AL and BLEU of each model.
The length was calculated as the number of subwords in a segment, and the previous segment was concatenated to the next segment  when the previous segment has no translation output.

Segmentation with fixed size 16 has some segments shorter than size 16 because the sentence length is not always a multiple of 16.

Compared with  ICLP model, Meaningful Unit has wider distribution, and the most segments consist of two subwords. These short segments have less context information and can output longer segment translation than expected. This would be one of the reason why our proposed method outperformed Meaningful Unit.

\subsection{Controlling latency}

\begin{figure}[t]
 \centering
 \centerline{\includegraphics[width=8.5cm]{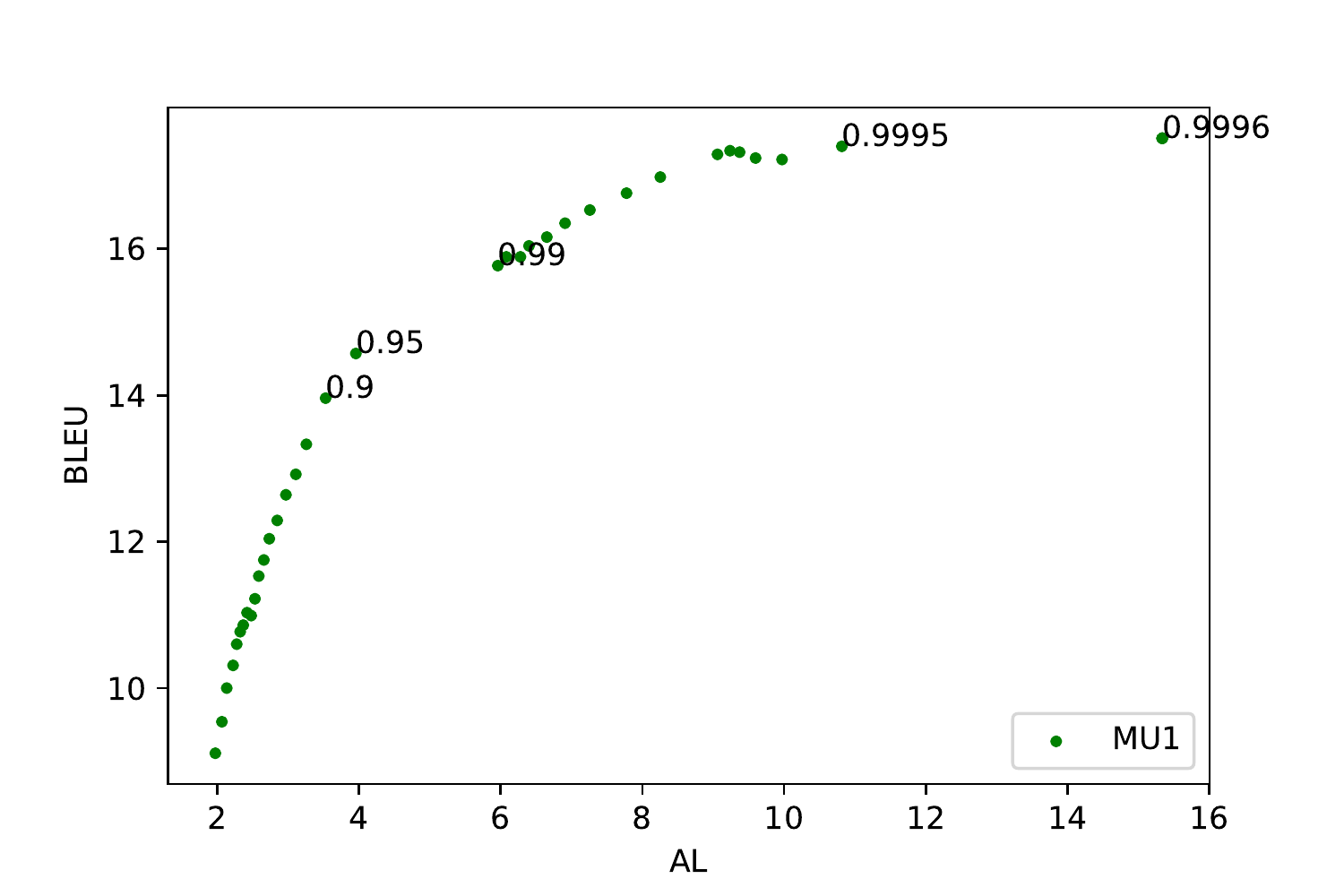}}
 \caption{BLEU and AL with different chunk segmentation thresholds for Meaningful Unit}
\label{fig:control_MU}
\end{figure}
\begin{figure}[t]
 \centering
 \centerline{\includegraphics[scale=0.55]{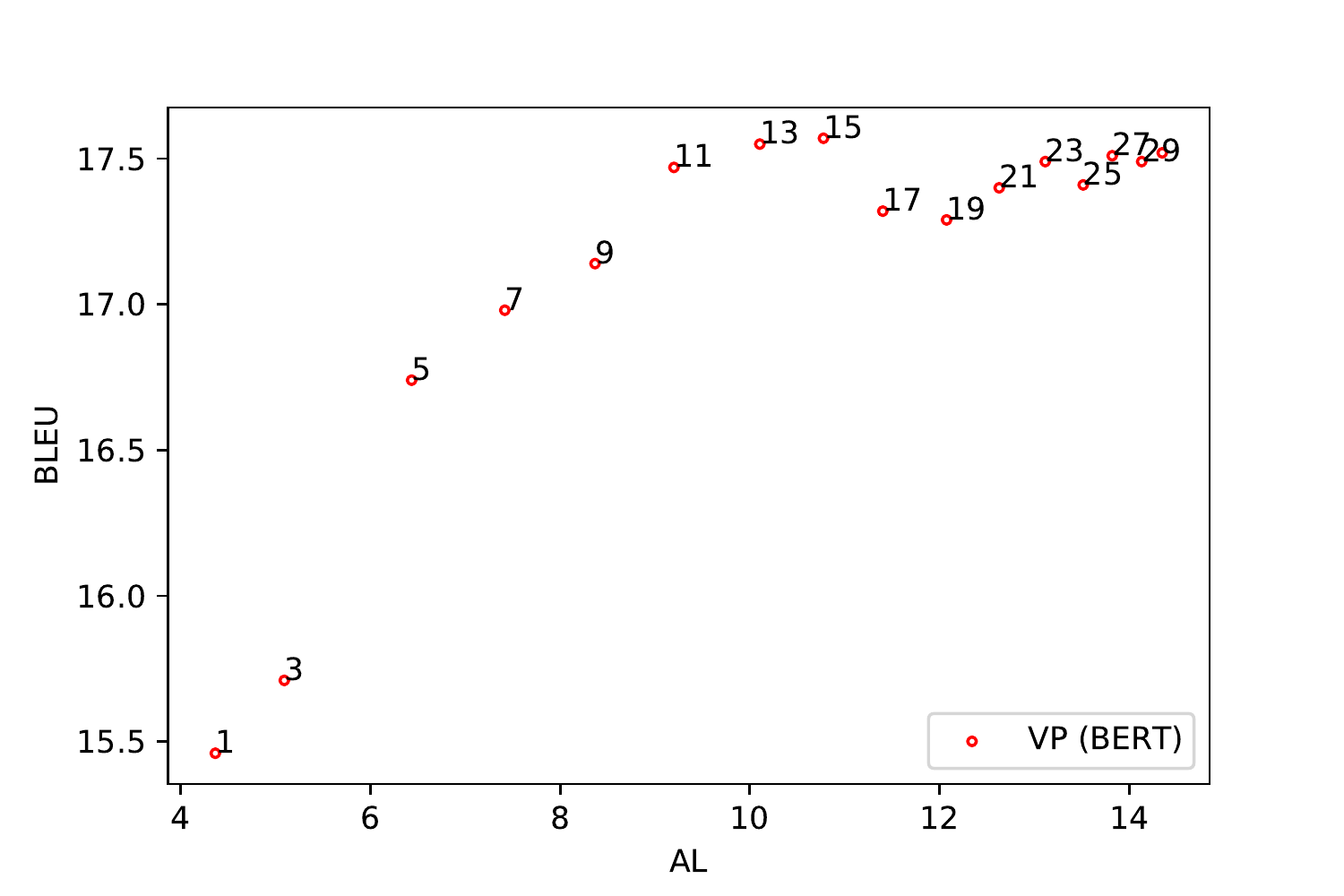}}
 \caption{BLEU and AL with different chunk size thresholds for the proposed method}
\label{fig:control_VP}
\end{figure}

In Figures~\ref{fig:control_MU} and \ref{fig:control_VP}, each plot is labeled by the corresponding value of the hyperparameter of inference. It is difficult to control latency for Meaningful Unit as shown in the figure. BLEU scores of hyperparameters from 0.9996 to 0.9999 were also the same as that of a full-sentence translation model.

In contrast, our proposed method can easily control latency because it uses the minimum chunk length as an intuitive hyperparameter to adjust it.

\subsection{How many words to wait}
Compared with the fixed-size segmentation model, our proposed model and Meaningful Unit have  a disadvantage in AL, which is caused by the look-ahead approach. Despite this disadvantage, our proposed approach outperformed the fixed-size segmentation in a wide range of AL. This means the benefit of looking at the future words and finding a better boundary outweighed the above disadvantage. 

\subsection{Performance of ICLP}

\begin{table}[t]
\centering
\begin{tabular}{lccc}
\hline
\textbf{Label}  & \textbf{Precision} & \textbf{Recall} & \textbf{F1}  \\
\hline
\textbf{NP} & 0.90 & 0.94 & 0.92\\
\textbf{VP} & 0.89 & 0.97 & 0.93\\
\textbf{NN} & 0.95 & 0.97 & 0.96\\
\textbf{,} & 0.98 & 1.00 & 0.99\\
\textbf{PP} & 0.85 & 0.93 & 0.89\\
\textbf{S} & 0.87 & 0.52 & 0.65\\
\hline
\end{tabular}
\caption{Results of label prediction (BERT)}
\label{tb:bertresult}
\end{table}

\begin{table}[!t]
\centering
\begin{tabular}{lccc}
\hline
\textbf{Label}  & \textbf{Precision} & \textbf{Recall} & \textbf{F1}  \\
\hline
\textbf{NP} & 0.85 & 0.89 &0.87 \\
\textbf{VP} &0.91 & 0.94 & 0.92 \\
\textbf{NN} &0.93 & 0.92 & 0.92\\
\textbf{,} &0.98 &1.00 &0.99 \\
\textbf{PP} &0.78 &0.94 &0.86\\
\textbf{S} &0.84 &0.52 &0.64\\
\hline
\end{tabular}
\caption{Results of label prediction (LSTM)}
\label{tb:lstmresult}
\end{table}

\begin{figure}[t]
 \centering
 \centerline{\includegraphics[width=8.5cm]{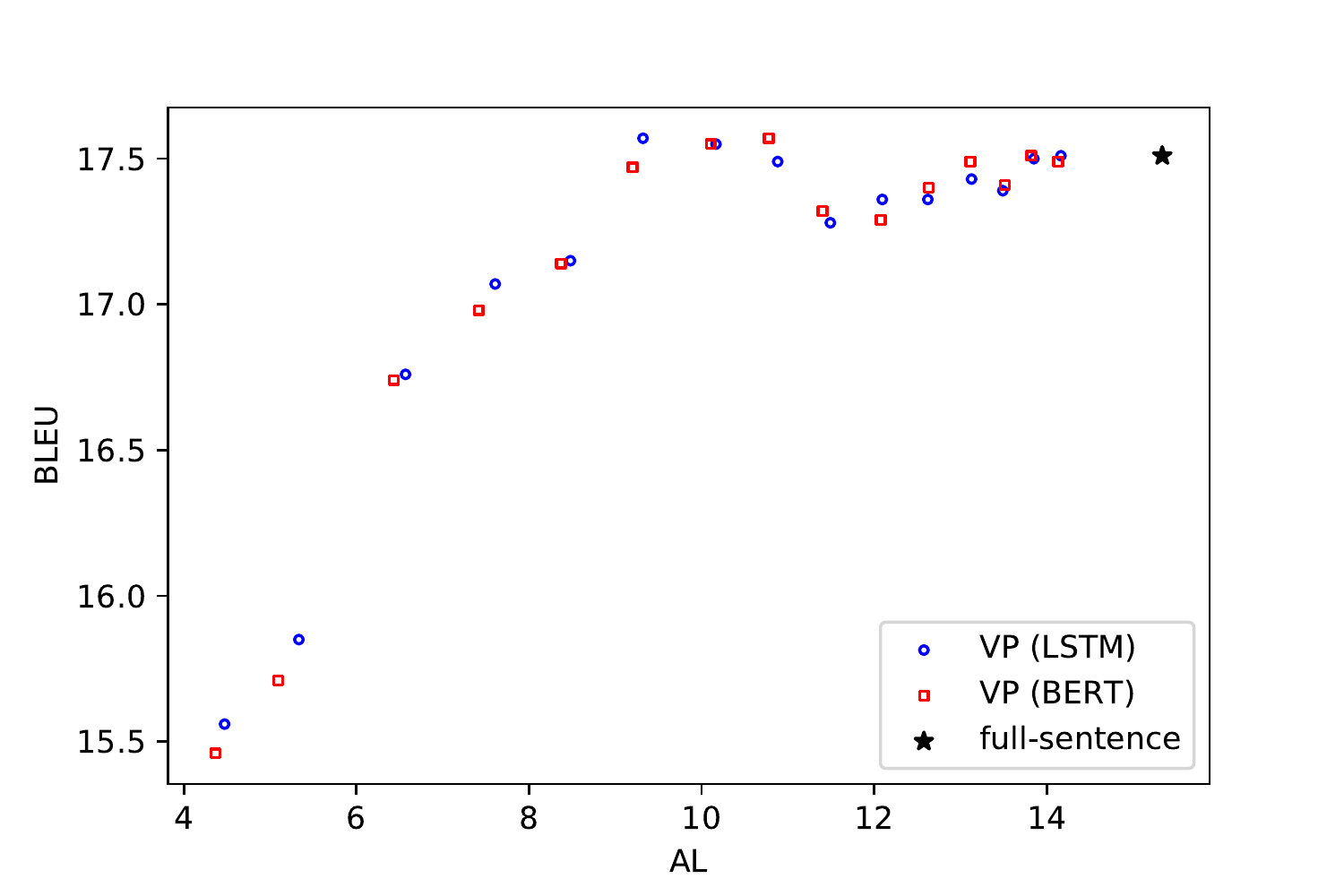}}
 \caption{Comparison between the use of LSTM- and BERT-based ICLP}
\label{fig:result_lstm}
\end{figure}

\begin{table}[!t]
\centering
\begin{tabular}{lccc}
\hline
\textbf{Label}  & \textbf{Precision} & \textbf{Recall} & \textbf{F1}  \\
\hline
\textbf{NP} &0.62 &0.85 &0.72 \\
\textbf{VP} &0.75 &0.80 &0.78 \\
\textbf{NN} &0.60 &0.78 &0.68\\
\textbf{,} &0.41 &0.34 &0.37 \\
\textbf{PP} &0.50 &0.47 &0.48\\
\textbf{S} &0.77 &0.62 &0.69\\
\hline
\end{tabular}
\caption{Results of label prediction (BERT) without look-ahead}
\label{tb:nolookahead}
\end{table}

Tables~\ref{tb:bertresult} and \ref{tb:lstmresult} show the results in precision and recall of the one look-ahead ICLP models.
The LSTM-based ICLP was better in precision, but the BERT-based ICLP was better in recall for \texttt{VP}.
Figure~\ref{fig:result_lstm} compares them in the downstream simultaneous translation.
The lines connected by dots nearly overlapped, so there was no large difference in BLEU score.
LSTM is more efficient than BERT in incremental processes, so it is suitable for practical usage.

Table~\ref{tb:nolookahead} shows the results by the ICLP model without one look-ahead approach. 
Compared with Table~\ref{tb:bertresult}, the scores are much lower. 
One look-ahead approach was important to improve its performance.

\subsection{En-De translation}
\begin{figure}[!t]
 \centering
 \centerline{\includegraphics[width=8.5cm]{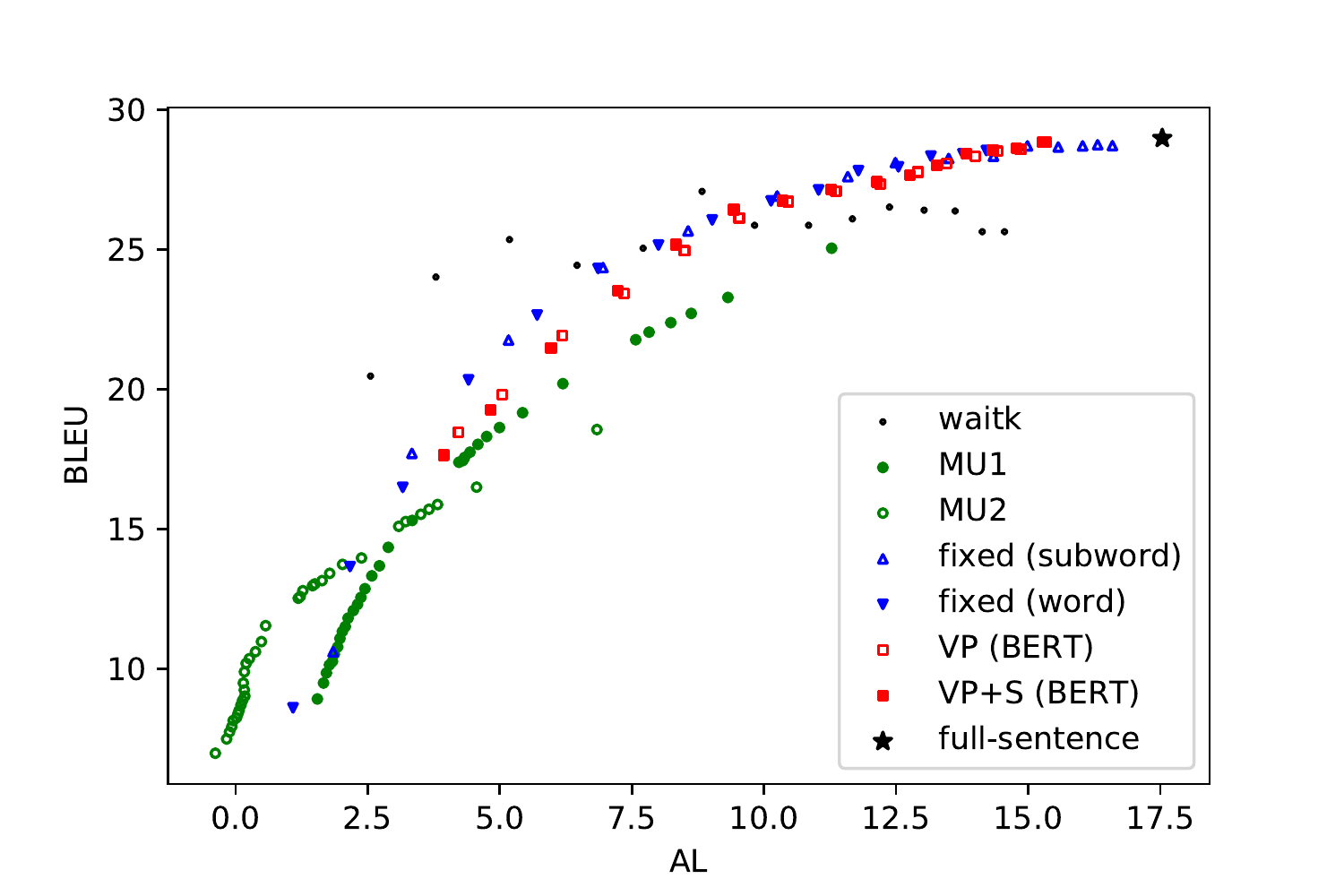}}
 \caption{Scatter plot of BLEU and AL (En-De)}
\label{fig:result_en_de}
\end{figure}

We conducted additional experiments in En-De translation to investigate the performance in a different language.
German is another language with different word order from English especially in verbs and also suffers from the reordering problem.
Figure~\ref{fig:result_en_de} shows the results.
This is almost the opposite of the results of the En-Ja translation.
The proposed boundary decision rules used for En-Ja translation were not so effective for En-De translation, so we need to find other rules to detect boundaries in En-De translation.

\section{Conclusion}
We proposed a novel segmentation method for simultaneous translation that uses simple rules and ICLP. Our proposed method is simple, and it outperformed the baselines in the quality-latency trade-off in En-Ja translation. On the other hand, the proposed method did not work effectively in En-De translation due to the smaller word order differences than those in En-Ja translation.

In future work, we expect to extract segmentation rules automatically and apply these rules to other language pairs as well. 

\section{Acknowledgements}
Part of this work was supported by JSPS KAKENHI Grant Numbers JP21H05054 and JP21H03500.

\bibliography{anthology,custom}
\bibliographystyle{acl_natbib}

\appendix



\end{document}